\documentclass[11pt,a4paper]{article}
\usepackage[hyperref]{emnlp2020}

\usepackage{times}
\usepackage{latexsym}
\usepackage{graphicx}
\usepackage{bm}
\usepackage{verbatimbox}
\usepackage{booktabs}
\usepackage{amsmath}
\usepackage{multirow}
\usepackage{paralist}
\usepackage{enumitem}
\usepackage{xcolor}
\usepackage{amsfonts}

\DeclareMathOperator*{\softmax}{softmax}
\DeclareMathOperator*{\MLP}{MLP}
\DeclareMathOperator*{\Transformers}{Transformers}

\usepackage{microtype}

\aclfinalcopy 
\def\aclpaperid{1191} 

\title{Towards Interpretable Reasoning over Paragraph Effects in Situation}

\author{
Mucheng Ren$^{1*\dag}$
        \ Xiubo Geng$^{2\dag}$,
        \ Tao Qin\textsuperscript{3},
        \ Heyan Huang\textsuperscript{1}\and
        Daxin Jiang$^{2\ddag}$ \\
\textsuperscript{1}School of Computer Science and Technology, Beijing Institute of Technology, Beijing, China
 \\
\textsuperscript{2}STCA NLP Group, Microsoft\\
\textsuperscript{3}Microsoft Research Asia \\
\texttt{\{renm,hhy63\}@bit.edu.cn} \\
\texttt{\{xigeng,taoqin,djiang\}@microsoft.com}
}

\date{}

\begin{document}
\maketitle
\renewcommand{\thefootnote}{\fnsymbol{footnote}}
\footnotetext[1]{Work done during internship at STCA NLP Group, Microsoft.}
\footnotetext[2]{Equal Contribution}
\footnotetext[3]{Corresponding author}
\renewcommand{\thefootnote}{\arabic{footnote}}

\begin{abstract}
We focus on the task of reasoning over paragraph effects in situation, which requires a model to understand the cause and effect described in a background paragraph, and apply the knowledge to a novel situation. Existing works ignore the complicated reasoning process and solve it with a one-step ``black box" model. Inspired by human cognitive processes, in this paper we propose a sequential approach for this task which explicitly models each step of the reasoning process with neural network modules. In particular, five reasoning modules are designed and learned in an end-to-end manner, which leads to a more interpretable model. Experimental results on the ROPES dataset demonstrate the effectiveness and explainability of our proposed approach.  
\end{abstract}

\section{Introduction}
As a long-standing fundamental task of natural language processing, machine reading comprehension (MRC) has attracted remarkable attention recently and different MRC datasets have been studied~\citep{rajpurkar-etal-2018-know,dua-etal-2019-drop,choi-etal-2018-quac,yang-etal-2018-hotpotqa}, among which reasoning over paragraph effects in situation (ROPES for short) is a very challenging scenario that needs to understand knowledge from a background paragraph and apply it to answer questions in a novel situation. Table~\ref{tab:example} shows an example of the ROPES dataset~\citep{lin-etal-2019-reasoning}, where the background passage states that developmental difficulties could usually be treated by using iodized salt, the situation passage describes two villages using different salt, and questions about which village having more/less people experiencing developmental difficulties need to be answered. 

\par
\begin{table}[t!]
{\begin{tabular}[c]{|p{0.95\linewidth}|}
\hline
\textbf{Background}
\\
Before \textcolor{orange}{iodized salt} was developed, some people experienced a number of \textcolor{blue}{developmental difficulties}, including problems with thyroid gland function and mental retardation. In the 1920s, we learned that these conditions could usually be treated easily with the addition of iodide anion to the diet. One easy way to increase iodide intake was to add the anion to table salt.
\\ 
\textbf{Situation}\\
People from two villages ate lots of salt. People from {\color[HTML]{009901} Salt village} used \textcolor{orange}{regular salt}, while people from {\color[HTML]{009901} Sand village} people used \textcolor{orange}{iodized salt} in their diets, after talking to specialists.
\\ 
\textbf{Q\&A}\\ 
Q: Which village had more people experience \textcolor{blue}{developmental difficulties}?   A: Salt \\
Q: Which village had less people experience \textcolor{blue}{developmental difficulties}?   A: Sand \\
\hline
\end{tabular}}
\caption{An example from the ROPES dataset. Effect property tokens are highlighted in \textcolor{blue}{blue}, cause property tokens in \textcolor{orange}{orange}, and world tokens in {\color[HTML]{009901} green.}}
\label{tab:example}
\end{table}

Almost all existing works \cite{lin-etal-2019-reasoning, khashabi2020unifiedqa,dua2019orb,gardner2020evaluating} for this task adopt a standard MRC approach based on deep learning in one step: the question and a pseudo passage constructed by concatenating the background and situation are fed into a large pre-trained model (e.g. RoBERTa large), and the answer is predicted directly by the model. However, the ROPES task is more complicated than traditional MRC since it requires a model to not only understand the causes and effects described in a background paragraph, but also apply the knowledge to a novel situation. Ignoring the understanding and reasoning process hinders such models from achieving their best performance. Consequently, the best F1 (61.6\%) achieved so far is far below human performance (89.0\%). More importantly, such a one-step approach makes the reasoning process unexplainable, which is of great importance for complicated reasoning tasks. 

We observe that human solve this kind of complicated reasoning tasks in a sequential manner with multiple steps~\cite{evans1984heuristic,sloman1996empirical,sheorey2001differences,mokhtari2002assessing,mokhtari2002measuring}. As shown in Table \ref{tab:example}, the background paragraph usually states the relationship between a cause property and an effect property, the situation describes multiple worlds each of which is associated with a specific value in terms of the cause property. Human usually does reasoning in a multi-step process: (1) identifying mentioned worlds, (2) identifying the cause and effect property, (3) understanding the relationship between the cause and effect property, (4) comparing identified worlds in terms of the cause property, and (5) reasoning about the comparison of mentioned worlds in terms of the effect property based on (3) and (4). 

Inspired by human cognitive processes, in this paper, we propose a sequential approach that leverages neural network modules to implement each step of the above process\footnote{The code is publicly available at \url{https://github.com/Borororo/interpretable_ropes}.}. Specifically, we define

\begin{itemize}
    \item a \textit{World Detection} module to identify potential worlds,
    \item an \textit{Effect and Cause Detection} module to identify effect and cause property,
    \item a \textit{Relation Classification} module to understand the relationship between effect and cause,
    \item a \textit{Comparison} module to compare identified worlds in terms of the cause property, and
    \item a \textit{Reasoning} module to infer comparison of mentioned worlds in terms of the effect property.
\end{itemize}

These modules are trained in an end-to-end manner, and auxiliary loss over intermediate latent decisions further boosts the model accuracy.

Explicitly modeling the sequential reasoning process has two advantages. First, it achieves better performance since the complicated reasoning process is decomposed into more manageable sub-tasks and each module only needs to focus on a simple sub-task. Second, intermediate outputs provide a better understanding of the reasoning process, making the learnt model more explainable.

Experimental results on the ROPES dataset demonstrate the effectiveness and explainability of our proposed approach. It surpasses the state-of-the-art model by a large margin (6\% absolute difference) in the five-fold cross-validation setting. Furthermore, analyses on intermediate outputs show that each module in our learnt model performs well on its corresponding sub-task and well explains the reasoning process.

\section{Related Work}
Neural network modules have been studied by several works.~\citet{andreas2016neural} propose neural module networks with a semantic parser on visual question answering.~\citet{jiang-bansal-2019-self} apply a self-assembling modular network with only three modules: Find, Relocate and Compare to Hotpot QA~\citep{yang-etal-2018-hotpotqa}. ~\citet{gupta2019neural} extend the neural module networks to answer compositional questions against a paragraphs of text as context, and perform symbolic reasoning on the self-pruned subset of DROPS~\citep{dua-etal-2019-drop}. Compared with them, we focus on a more challenging MRC task: reasoning over paragraph effects in situation, which has been rarely investigated and needs more complex reasoning. So far as we know, the only two works (i.e. ~\citep{lin-etal-2019-reasoning} and~\citep{khashabi2020unifiedqa}) on this topic uses a one-step ``black box" model. Such an approach performs well on some questions at the expense of limited intepretability. Our work solves this task in a logical manner and exposes intermediate reasoning steps which improves performance and interpretability concurrently.\\

\begin{figure*}[ht]
    \centering
    \includegraphics[height=0.29\textheight,width=\textwidth]{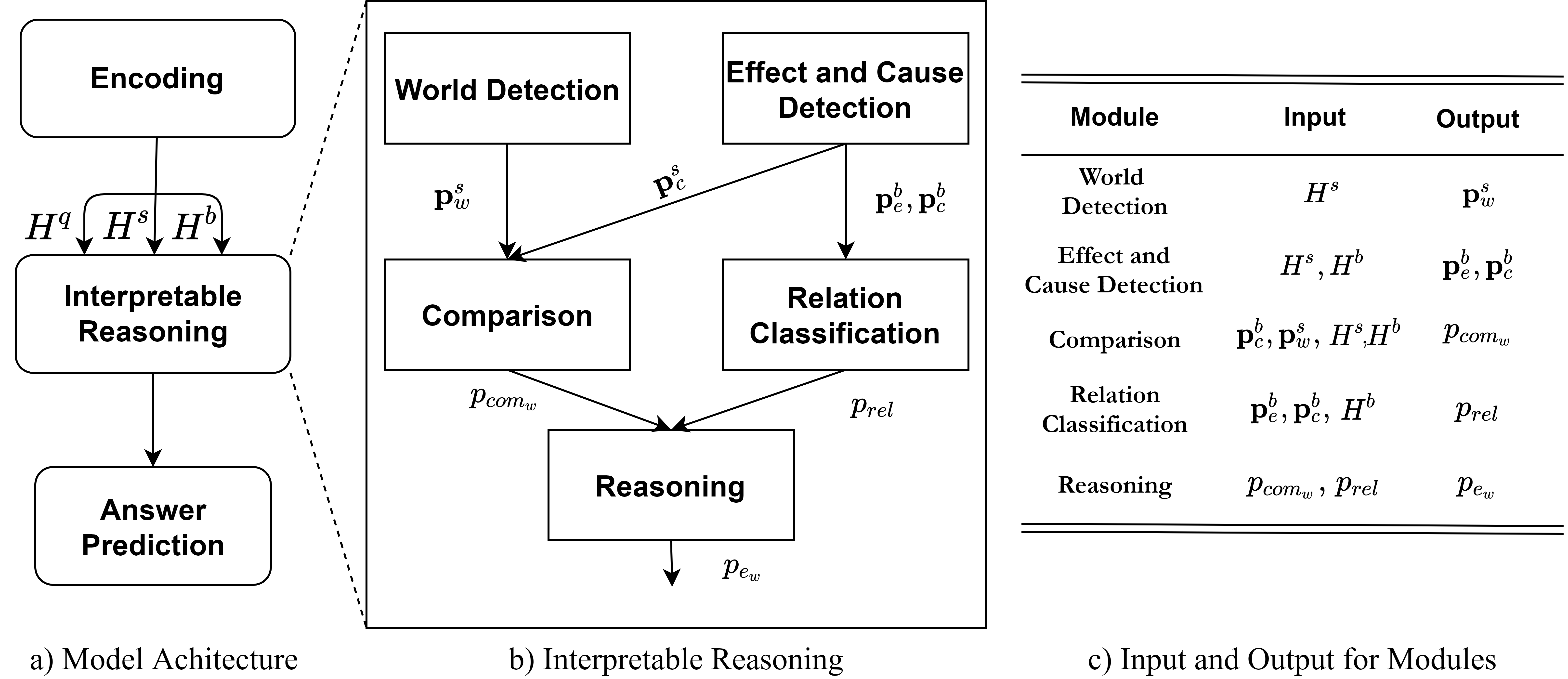}
    \caption{The left part is the architecture of our model. The middle part is the interpretable reasoning component in our model. The right part is the summary for inputs and outputs flowing between each module. The encoded contextual representations, $\bm{H^q,H^s,H^b}$, serve as global variables for the interpretable reasoning component.}
    \label{fig:IR}
\end{figure*}
\section{Methodology}
\label{sec:methods}
As shown in Figure~\ref{fig:IR}, our approach consists of three components which are contextual encoding, interpretable reasoning, and answer prediction. 

\subsection{Contextual Encoding}
We use RoBERTa~\citep{devlin-etal-2019-bert,liu2019roberta} to encode background, situation and question together and generate contextualized embeddings. Specifically, given a background passage $B =\left \{ b_{i} \right \}_{i=1}^{m}$, a situation passage $S =\left \{ s_{j} \right \}_{j=1}^{n}$ and a question $Q =\left \{ q_{k} \right \}_{k=1}^{l}$, we concatenate them with special tokens as $\left \langle s \right \rangle q_1,\dots, q_l \left \langle \slash s \right \rangle \left \langle \slash s \right \rangle s_1,\dots,s_n;b_1,\dots,b_m \left \langle \slash s \right \rangle$, which is then fed into a series of successive transformer blocks contained in RoBERTa, 
\begin{equation}
    \bm{H^q,H^s,H^b}= \Transformers(Q,S,B),
\end{equation}
where $\bm{H^b} \in \mathbb{R}^{m \times d}$, $\bm{H^s} \in \mathbb{R}^{n \times d}$, and $\bm{H^q} \in \mathbb{R}^{l \times d}$ are contextual embeddings for the background, situation, and question, respectively, $d$ is the dimension for hidden states. 

\subsection{Interpretable Reasoning}
\subsubsection*{World Detection}
The module aims to identify concerned worlds from situation according to a question. Take Table \ref{tab:example} as an example, the question cares about two worlds, \textit{Sand Village} and \textit{Salt Village}. To achieve that, we apply a multilayer perceptron (MLP) over the situation representations $\bm{H^s}$ and normalize the projected logits (using a softmax function) to get attention over all situation tokens for each world,
\begin{align} 
 \bm{p_{w_1}^{s}}&=\softmax(\MLP(\bm{H^s};\theta_{w_1}))\in\mathbb{R}^{n}, \\
 \bm{p_{w_2}^{s}}&=\softmax(\MLP(\bm{H^s};\theta_{w_2}))\in\mathbb{R}^{n},
\end{align}
where $\bm{p_{w_1}^\bm{s}}$ and $\bm{p_{w_2}^{s}}$ are the attention vectors over situation for the first and second world, $\theta$'s are learnable parameters of MLP. Note that since most examples in the ROPES dataset are related to two concerned worlds, we identify two worlds in our model. However, we can handle multiple worlds by simply extending the module with more MLPs. 

\subsubsection*{Effect and Cause Detection}
This module aims to identify effect and cause properties described in the background. To achieve that, another MLP is used to identify the effect property,
\begin{gather}
 \bm{p_{e}^{b}}=\softmax(\MLP(\bm{H^b};\theta_{e}))\in\mathbb{R}^{m}.
\end{gather}
Here $\bm{p_{e}^b}$ is the attention vector over background tokens in terms of the effect property, which attends more to tokens of effect property. Take Table \ref{tab:example} as an example, $\bm{p_e^b}$ is the attention over background tokens, whose value is much larger for \textit{developmental difficulties} than other tokens. 

Next, we apply a \textit{relocate} operation which re-attends to the background based on the situation and is used to find the cause property in the background (e.g., shifting the attention from \textit{developmental difficulties} to \textit{iodized salt} in Table \ref{tab:example}). This is achieved with the help of a situation-aware background-to-background attention matrix $\bm{R}\in\mathbb{R}^{m \times m}$,
\begin{gather}
        \bm{R}_{ij} = {\bm{w_{relo}}}^{T} \left [(\bm{s} +\bm{H_{i}^{b}});\bm{H_{j}^{b}};(\bm{s}+\bm{H_{i}^{b}})\odot \bm{H_{j}^{b}} \right ], \\
    \bm{s} = \frac{1}{n}\sum_{i}^{n}\bm{H_{i}^{s}}\in \mathbb{R}^{d},
\end{gather}
where [;] denotes the concatenation operation and $\odot$ is Hadamard product. $\bm{w_{relo}} \in \mathbb{R}^{3d}$ is a learnable parameter vector, $\bm{s}$ can be viewed as an embedding of the whole situation. Then each row of $\bm{R}$ is normalized using the softmax operation. Finally we get the attention vector over background tokens in terms of the cause property $\bm{p_c^b}$,
\begin{gather}
    \bm{p_{c}^{b}}=\bm{R}^T\bm{p_{e}^{b}} \in \mathbb{R}^m.
\end{gather}
Here $\bm{p_c^b}$ should attend more to the tokens of effect property. For example, \textit{iodized salt} will get larger attention value than other tokens in the background.

\subsubsection*{Relation Classification}
This module aims to predict the qualitative relation between effect and cause property. Take Table \ref{tab:example} as an example, the cause property \textit{iodized salt} and the effect property \textit{developmental difficulty} is negatively correlated. To achieve that, we first derive and concatenate representations of cause and effect property by averaging background representation $\bm{H^b}$ weighted by according attention vector, $\bm{p_c^b}$ and $\bm{p_e^b}$. Next, we adopt another MLP stacked with softmax to get corresponding probabilities,
\begin{gather}
    \bm{p_{rel}} =  \softmax(\MLP((\bm{{H^b}}^T\bm{p_{e}^{b}};\bm{{H^b}}^T\bm{p_{c}^{b}});\theta_{rel})),
\end{gather}
where $\bm{p_{rel}} = [p_{rel-}$, $p_{rel+}]$ denotes probability of negative and positive relation, $\theta_{rel}$ is a learnable parameter in the MLP. In the example shown in Table \ref{tab:example}, $p_{rel-}$ is supposed to be larger than $p_{rel+}$.

\subsubsection*{Comparison}
This module aims to compare the worlds in terms of the cause property. For example, world 1 (\textit{salt village}) is more relevant to \textit{iodized salt} than world 2 (\textit{sand village}) in Table \ref{tab:example} since people in \textit{salt village} use \textit{iodized salt} while people in \textit{sand village} use \textit{regular salt}. 

This is achieved by three steps. First, we derive the attention of cause property over situation $\bm{p_c^s}$ from $\bm{p_c^b}$ with a similarity matrix $\bm{M}\in\mathbb{R}^{n\times m} $ between situation and background, 
\begin{gather}
    \bm{M}_{ij} ={\bm{H_{i}^{s}}}\bm{W}_{sb}{\bm{{H_{j}^b}}^T}, \\
    \bm{p_{c}^{s}}=  \bm{M}\bm{p_{c}^{b}} \in \mathbb{R}^n,
\end{gather}
where $\bm{W}_{sb} \in \mathbb{R}^{d\times d} $ are learnable parameters. 

Second, we use $\bm{p_{w}^{s}}$ to mask out irrelevant cause property for each world. This part ensures the alignment between each world and its cause property, which is critical when one situation contains multiple worlds.
\begin{gather}
    \bm{p_{c_{w_1}}^{s}}= \softmax(\bm{p_{w_1}^{s}}\odot \bm{p_{c}^{s}}), \\
    \bm{p_{c_{w_2}}^{s}}= \softmax(\bm{p_{w_2}^{s}} \odot \bm{p_{c}^{s}}).
\end{gather}
\par

Third, each world's cause property is evaluated by a bilinear function in terms of its relevance to the cause property in background, which is further normalized into a probability with softmax,
\begin{gather}
    logit_{w_1} = (\bm{{H^b}}^T\bm{p_{c}^{b}})^{T}\bm{W}_{com}(\bm{{H^s}}^T\bm{p_{c_{w_1}}^{s}}), \\
    logit_{w_2} = (\bm{{H^b}}^T\bm{p_{c}^{b}})^{T}\bm{W}_{com}(\bm{{H^s}}^T\bm{p_{c_{w_2}}^{s}}), \\
    \bm{p_{com_w}} = \softmax(logit_{w_1},logit_{w_2}),
\end{gather}
\par
where $\bm{W_{com}}\in\mathbb{R}^{d\times d}$ is a learnable matrix, $\bm{{H^b}}^T\bm{p_{c}^{b}}$ represents expected embedding of cause property in background, $\bm{{H^s}}^T\bm{p_{c_{w_i}}^{s}}$ represents expected embedding of cause property for world $i$, $p_{com_{w_i}}$ denotes the probability that world $i$ is relevant to cause property. 
\subsubsection*{Reasoning}
Given the relationship between effect property and cause property, $p_{rel+}$ and $p_{rel-}$, and the comparison between worlds in terms of cause property $p_{com_{w_i}}$, this module infers comparison between identified worlds in terms of the effect property. Take Table \ref{tab:example} as an example, given the negative relationship between \textit{developmental difficulties} and \textit{iodized salt}, and \textit{salt village} uses more \textit{iodized salt} than \textit{sand village}, we infer that people in \textit{sand village} are more likely to have \textit{developmental difficulties}.

To this end, we have
\begin{gather}
    p_{e_{w_1}}=  p_{com_{w_1}}\times p_{rel+} + p_{com_{w_2}}\times p_{rel-}, \\
    p_{e_{w_2}}=  p_{com_{w_1}}\times p_{rel-} + p_{com_{w_2}}\times p_{rel+},
\end{gather}
where $p_{e_{w_i}}$ is the probability that world $i$ is more relevant to effect property. 
\par

\subsection{Answer Prediction} \label{sec:ans}
Given intermediate outputs from the interpretable reasoning component, this module predicts the final answer for a question. Specifically, we first convert these intermediate outputs into text spans or 0/1 class as follow.
\begin{itemize}
    \item We take two steps to convert an attention vector output by \textit{World Detection} or \textit{ Effect and Cause Detection} into a text span. First, the token with the highest probability is selected. Then it is expanded with left and right neighbors which are continuous spans and the probability of each token is larger than threshold $t$. In our experiment we set $t=\frac{1}{l}$, where $l$ is the length of the paragraph.
    \item For \textit{Comparison, Relation Classification, and Reasoning}, we select the class with the highest probability. 
\end{itemize}

Then we synthesize a sentence $\hat{s}$ in the format of \texttt{[World 1] has [larger/smaller] [Effect Property] than [World 2]}, where choosing “larger” or “smaller” depends on results from the \textit{Reasoning} module. Take Table \ref{tab:example} as an example, the synthetic sentence is \textit{Salt village has larger developmental difficulties than Sand village}. Such synthetic text explicitly expresses comparison between the identified worlds in terms of the effect property. Finally, we concatenate it with the situation $s$ and question $q$ as $\left \langle s \right \rangle q;s \left \langle \slash s \right \rangle\left \langle \slash s \right \rangle \hat{s} \left \langle \slash s \right \rangle$, and feed them into RoBERTa which directly predicts the starting and end position of the final answer.

\subsection{Model Training}
\label{sec:loss}
Two models (i.e. interpretable reasoning model; and answer prediction model) are learned in our approach. 

\paragraph{Interpretable Reasoning}
The final loss function for interpretable reasoning is defined as
\begin{gather}
    l_{intp} =  -\sum_{\bm{x}\in \bm{X}} \alpha_{\bm{x}}\bm{\Tilde{x}}^{T}log(\bm{x}).
\end{gather}
Here $\bm{X} = \{\bm{p_{w_1}^{s}}\in\mathbb{R}^{n},\bm{p_{w_2}^{s}}\in\mathbb{R}^{n},\bm{p_{e}^{b}}\in\mathbb{R}^{m},\bm{p_{c}^{b}}\in\mathbb{R}^{m},\bm{p_{c_{w_1}}^{s}}\in\mathbb{R}^{n},\bm{p_{c_{w_2}}^{s}}\in\mathbb{R}^{n}, \bm{p_{rel}}\in\mathbb{R}^{2},\bm{p_{com_{w}}}\in\mathbb{R}^{2},\bm{p_{e_{w}}}\in\mathbb{R}^{2} \} $ are predictions of different modules, $\bm{\Tilde{x}}^{T}\in \left \{ 0,1 \right \}^n$ or $ \bm{\Tilde{x}}^{T}\in\left \{0,1 \right \}^m$ or $ \bm{\Tilde{x}}^{T}\in\left \{0,1 \right \}^2$ are corresponding gold labels, and $\alpha_{x}$ is the weight for module $x$.

\paragraph{Answer Prediction}
The training objective of the answer prediction model is defined as 
\begin{gather}
    l_{ans} = -( \bm{\Tilde{s}}\log(\bm{s})+\bm{\Tilde{e}}\log(\bm{e})),
\end{gather}
where $\bm{s}, \bm{e} \in \mathbb{R}^{m+n+k}$ are predicted probabilities of the starting and end position, $k$ is the length of the synthetic sentence $\hat{s}$, and $\bm{\Tilde{s}},\bm{\Tilde{e}} \in \left \{0,1 \right \}^{m+n+k}$ are corresponding gold labels. 

\section{Experimental Setup}
\subsection{Dataset}
We evaluate our proposed approach on the ROPES \cite{lin-etal-2019-reasoning} dataset\footnote{\url{https://leaderboard.allenai.org/ropes/submissions/get-started}}. So far as we know, it is the only dataset that requires reasoning over paragraph effects in situation. Given a background paragraph that contains knowledge about relations of causes and effects and a novel situation, questions about applying the knowledge to the novel situation need to be answered. Table \ref{tab:example} shows an example and Table~\ref{tab:statistics} presents the statistics. To be noticed, different from other extractive MRC datasets, train/dev/test set in ROPES is split based on annotators instead of context~\citep{geva2019we,lin-etal-2019-reasoning}. This might pose a large data bias in each set. For example, as can be seen in Table~\ref{tab:statistics}, dev and test sets have similar numbers of questions, while the vocabulary size of background and situation in test set is $2\times$ and $2.7\times$ larger than that in dev set. The same thing happens on the size of question vocabulary, which indicates the existence of the distribution gap between train/dev and test sets and it might lead to underestimate/overestimate the performance of a model.
\begin{table}[t!]
\resizebox{\linewidth}{!}{%
\begin{tabular}{@{}lccc@{}}
\toprule[2pt]
\textbf{Statistics }                & \textbf{Train} & \textbf{Dev}  & \textbf{Test}  \\ \midrule
background vocabulary size & 8,616  & 2,008 & 3,988  \\
situation vocabulary size  & 6,949  & 1,077 & 2,736  \\
question vocabulary size   & 1,457  & 1,411 & 1,885  \\ \midrule
avg. background length     & 121.6 & 90.7 & 123.1 \\
avg. situation length      & 49.1  & 63.4 & 55.6  \\
avg. question length       & 10.9  & 12.4 & 10.6  \\ \midrule
No. of questions            & 10,924 & 1,688 & 1,710  \\
No. of annotators          & 7     & 2    & 2     \\ 
\bottomrule[2pt]
\end{tabular}%
}
\caption{ROPES statistics}
\label{tab:statistics}
\end{table}
\paragraph{Cross Validation} Because of the limited size of the official dev set and potential data bias between train/dev and test, we conduct 5-fold cross-validation to verify the effectiveness of the proposed approach. K-fold cross-validation assesses the predictive performance of the models and judges how they perform outside the sample to a new data set. Therefore it can assure unbiased results, avoid over-fitting, and testify the generalization capability of a model~\citep{burman1989comparative,browne2000cross,raschka2018model}. Specifically, we first exclude the labeled 1074 questions from the training data, and then split the remaining training plus dev data into five folds based on background, which ensures each subset has independent vocabulary space and they do not look through each other. For each split, we directly apply modules trained from auxiliary supervision data, and use the training data to train a model for answer prediction. Averaged results on the 5-fold cross-validation setting are reported.

\paragraph{Auxiliary Supervision} We randomly sampled 10\% (1,074 questions) of the original training data and labeled them for training the proposed modules. For each example, we label two concerned worlds, effect property in background and situation, values of cause property for two worlds in situation, comparison between two worlds in terms of cause property and effect property, and relationship between cause property and effect property. More detailed guidelines and labeled examples are in Appendix~\ref{sec:ASI}. Note the neural network modules are trained only on the labeled 1,074 questions.


\subsection{Implementation Details}
Our model is evaluated based on the pretrained language model RoBERTa large in Pytorch version\footnote{\url{https://github.com/huggingface/transformers}}. We train the five modules on one P100 16GB GPU and use four GPUs for predicting final answer. We tune the parameter $\alpha_x$'s according to the averaged performance of all modules, and set it to be 0.05 for span-based loss, 0.2 for the \textit{Comparison} and \textit{Relation} prediction, and 0.3 for the \textit{Reasoning} prediction. Evaluation metrics are EM and F1 which are same as the ones used in SQuAD\footnote{\url{https://github.com/huggingface/transformers/blob/master/src/transformers/data/metrics/squad_metrics.py}}. The detailed hyperparameters are described in the Appendix~\ref{sec:params}.

\subsection{Baseline}
We re-implemented the best model (RoBERTa) in the leaderboard\footnote{\url{https://leaderboard.allenai.org/ropes/submissions/public}} and achieved similar performance (Our implemented baseline achieves EM 69.0 / F1 71.1 on dev and EM 55.2 / F1 61.0 on test while the official one achieves EM 59.7 / F1 70.2 on dev and EM 55.4 / F1 61.1 on test). The basic idea is to leverage the RoBERTa large model~\citep{liu2019roberta} to encode the concatenation of question, background and situation, and then apply a linear layer to predict the starting and end position of an answer directly.

\section{Experimental Results}
\subsection{Question Answering Performance}
Table \ref{tab:cross} shows question answering performance of different models, where our approach outperforms the RoBERTa large model by 8.4\% and 6.4\% in terms of EM and F1 scores respectively. These results show that compared to one-step ``black box" model, our interpretable approach which mimics the human reasoning process has a better capability of conducting such complex reasoning.

Furthermore, we also list the performance of our approach and the baseline model when using only randomly sampled 10\% of training data in Table \ref{tab:cross}. That is, both the neural network modules and answer prediction model in our approach are trained with only 1074 questions. As seen in the table, our model learned from 10\% of training examples achieves competitive performance to the baseline model learned from full data (71.8\% v.s. 71.1\% in terms of F1 score). In contrast, the performance of the baseline model drops dramatically by 32\%. This indicates that traditional black-box approach requires much more training data while our approach has better generalization ability and can learn the reasoning capability with much fewer examples.

We also implement a rule-based answer prediction approach (detailed descriptions in the Appendix~\ref{sec:rules}), which are generated based on the same 10\% of training examples as in interpretable reasoning components. As shown in Table \ref{tab:cross} the rule-based approach performs worse than the RoBERTa model, indicating better generalization ability of pre-trained models. 

\begin{table}[!t]
\resizebox{\linewidth}{!}{%
\begin{tabular}{@{}lcccc@{}}
\toprule[2pt]
\multirow{2}{*}{\textbf{Model}} & \multicolumn{2}{c}{\textbf{Dev}} & \multicolumn{2}{c}{\textbf{Test}} \\ \cmidrule(l){2-5} 
 & EM         & F1         & EM          & F1  \\\midrule

RoBERTa$_{Large}^{*}$       & 61.4 & 68.4 & 64.0 & 71.1 \\
Ours & \textbf{73.0 }& \textbf{78.1} & \textbf{72.4} &\textbf{77.5} \\
\midrule
RoBERTa$_{Large}^{*}$ ($10\% $ data)        & - & - & 43.1 & 53.9  \\
Ours ($10\%$ data) & - & - & 60.9 & 71.8 \\
\midrule
Ours (rule-based)  & - & -& 54.3 & 65.5 \\

\bottomrule[2pt]
\end{tabular}%
}
\caption{Performance of different models on the ROPES dataset under cross-validation setting.}
\label{tab:cross}
\end{table}
\subsection{Case Study of Interpretability} \label{sec:explain}
The most remarkable difference between our model and the one-step ``black box" model is that our model outputs multiple intermediate predictions which well explains the reasoning process.
Note all modules in our model output probabilities or attention on input text, which are further fed into downstream ones for end-to-end learning. In order to explicitly visualize the output of each module, we take a similar approach to \S\ref{sec:ans} to convert these probabilities into a text span or a 0/1 classification.

We demonstrate the reasoning process of our model with a running example shown in Table \ref{tab:all}. Please see more examples in Appendix~\ref{sec:more_ex}. Here the background states the relationship between \textit{CPU load} and \textit{data volume}, i.e. \textit{CPU load goes up when processing larger volume of data}. The situation describes that Tory stored different sizes of data at a different time. For example, he stored \textit{301 Gigabytes} at \textit{8 AM} and went to \textit{sleep} at \textit{1 PM}. Finally, the question asks to compare CPU loads between 8 AM and 1 PM.
\begin{table}[t!]
{\begin{tabular}[c]{|p{0.95\linewidth}|}
\hline
\textbf{Background}
\\
\textcolor{orange}{Storing large volumes of data} - When storing XML to either file or database, the volume of data a system produces can often exceed reasonable limits, with a number of detriments: the access times go up as more data is read, \textcolor{blue}{CPU load goes up} as XML data takes more power to process, and storage costs go up....
\\ 
\textbf{Situation}\\
Tory had a busy day storing XML. At 7 AM, he stored 201 Gigabytes to the database. At {\color[HTML]{009901} 8 AM}, he stored \textcolor{orange}{301 Gigabytes} to the database. At 9 AM,... At 10 AM, ... At 11 AM, ... At 12 PM, ... At {\color[HTML]{009901} 1 PM}, he went to \textcolor{orange}{sleep} to finish storing XML later on that day.
\\ 
\textbf{Q\&A}\\ 
Q: What time did \textcolor{blue}{CPU load go up}: 8 AM or 1 PM?   A: 8 AM\\
\hline
\textbf{Predictions} \newline
\begin{tabular}{ll}
Worlds:& \textit{[8 AM,1 PM]} \\
Effect:&\textit{CPU load goes up} \\
$\text{Cause}^{B}$:&\textit{Storing large volumes of data }\\
$\text{Cause}_{\text{World1}}^{S}$:&\textit{301 Gigabytes}\\
$\text{Cause}_{\text{World2}}^{S}$:&\textit{sleep} \\
Cause Cmp: &\textit{World 1} \\
Relation: & \textit{positively related }\\
Effect Cmp: & \textit{World 1} \\
\textit{Final Answer}: & 8 AM 
\end{tabular}
\\
\hline
\end{tabular}}
\caption{A running example with visualized intermediate outputs of our approach.}
\label{tab:all}
\end{table}
As shown in Table \ref{tab:all}, our model outputs several intermediate results. First, it identifies two concerned worlds, \textit{8 AM} and \textit{1 PM} from the situation. Then it predicts the effect property, \textit{CPU load goes up}, given which the cause property in the background (i.e. \textit{storing large volumes of data}) and according values for the two worlds (i.e. \textit{301 Gigabytes} and \textit{sleep}) are predicted. Next, it compares the two worlds in terms of cause property and predicts that world 1 is larger than world 2. Also it predicts that the cause property and effect property is positively related, i.e. the relation is classified as 1. Finally, it reasons that world 1 takes higher CPU loads than world 2.
This example demonstrates that our approach not only predicts the final answer for the question, but also provides detailed explanations for the reasoning process. 
\begin{table}[t!]
\resizebox{\linewidth}{!}{%
\begin{tabular}{@{}lcclc@{}}
\toprule
          & \textbf{F1}  & \textbf{Fuzzy F1} &                                   & \textbf{Accuracy} \\ \midrule
World 1    & 83.5 &86.8 & Comparison                         & 83.8\%   \\
World 2  & 84.4 &86.1 & Relation                         & 84.5\%   \\
Effect& 67.8 &83.6 & Reasoning & 74.0\%   \\
$\text{Cause}^{B}$       & 57.6 &70.1 &  &   \\
$\text{Cause}_{\text{World1}}^{S}$ & 69.4 &81.3 &                                   &          \\
$\text{Cause}_{\text{World2}}^{S}$ & 58.2 &71.9 &                                   &          \\ \bottomrule
\end{tabular}%
}
\caption{Performance of Each Module}
\label{tab:Interpretable}
\end{table}
\subsection{Neural Network Module Performance}
\label{sec:modules}
Taking the same approach as in \S\ref{sec:explain}, we convert the output of each module into a text span or a predicted class. We manually sampled another 5\% of the training data, labeled them with outputs for each module, and evaluate the visualized results of all modules. Table~\ref{tab:Interpretable} summarizes the performance for each module, where the predicted text span is measured by F1 score and classification prediction is measured by accuracy.

\paragraph{World Detection} This module implements a similar capability as traditional extractive MRC, since both require to detect concerned text spans from a passage according to a question. Consequently, it achieves similar performance to top models of the popular SQuAD dataset\footnote{\url{https://rajpurkar.github.io/SQuAD-explorer/}}, where our \textit{World Detection} module reaches about 83\% F1 score and single RoBERTa large model on SQuAD gets about 89\%. The gap might come from different modeling styles. Our model predicts the probability of each token being concerned, while SQuAD models directly predict the starting and end position of an answer, which performs better on boundary detection. 

\paragraph{Effect and Cause Detection} Compared with \textit{World Detection}, the F1 score for this module decreases but actually still acceptable (F1=67.6 for $\text{Effect}^B$, 57.6 for $\text{Cause}^B$). The most possible reason is that, effects and causes are usually longer than world names. For example, the average length of world name is 1.2, while those of effect and cause are 2.7 and 2.2 respectively. Longer text span increases the difficulty of prediction. 

For above two span-related modules, we argue that since our model leverages the attention score in a soft way, it is less sensitive to accuracy of boundary changes. Therefore, we added another fuzzy F1 score for them. The fuzzy F1 of each question is set to 1 as long as its original F1 is larger than 0. As shown in Table \ref{tab:Interpretable}, the fuzzy F1 scores of these two modules increase to 70\%$\sim$86\%, indicating good reasoning capability of them.

\paragraph{Comparison, Relation Classification} These two modules essentially requires the capability of classification. The high accuracy (83.8\% and 84.5\%) indicates that our modeling approach can effectively leverages the prediction of upstream modules and does a good job on them.

\paragraph{Reasoning} Given the high accuracy of the \textit{Comparison} and \textit{Relation Classification} modules, the \textit{Reasoning} model achieves 74\% of accuracy, which provides high-quality input for final answer prediction.

\subsection{Error Analysis}
We randomly sampled 200 wrongly predicted questions to do error analysis and find that they fall into two major types, which are described below (Please see complete description of questions, backgrounds and situations in the Appendix~\ref{sec:ECM}).
\subsubsection{Type One Error}
Type one errors are caused by wrong model predictions, most of which occur in below three modules.
\paragraph{Wrong Predicted Worlds} Such errors are mainly caused by length imbalance between question and situation. Since situation is usually much longer than question, the \textit{World Detection} module might make the same predictions for different questions of same situation. 
\paragraph{Wrong Predicted Cause Property in Situation} Such errors are mainly caused by imbalanced descriptions for different worlds, where a situation describes details for one world but mentions another world with very few words. In such cases, the \textit{Comparison} module might assign the same cause property for different worlds in situation.
\paragraph{Wrong Predicted Comparison Results} Such errors often occur when two worlds are described with similar words, e.g. ``high" v.s. ``higher", or ``smoking" vs. ``not smoking", in which case the \textit{Comparison} module might be confused by similar expressions of two worlds and fail to compare them.  

\subsubsection{Type Two Error}
Type two errors occur when the proposed framework is not suitable to solve the questions. Here we list some example cases.

\paragraph{Missing Knowledge} Background paragraph does not provide sufficient knowledge for reasoning. For example, a background paragraph only describes information about fish while the questions asks fertilization take place inside or outside of mother's body for a mammal creature. 

\paragraph{Implicit Worlds} Concerned worlds in a question are not explicitly described in the situation. For example, a situation paragraph says that Mattew does intensive worksouts while the question asks that his strengths will increase or decrease when he stops working out. In such a case, the \textit{world} that Mattew stop working out is not explicitly described in the situation.

\paragraph{Additional Math Computation} Answering such questions requires additional math computation. For example, a background states the speed of sound waves in air/water/iron and the question asks how much faster a channel (with water) would be than another channel (with air). Answering such questions requires additional math computation (i.e. subtraction, addition etc.)

\section{Conclusion and Future Work}
In this paper, we aim to answer ROPES questions in an interpretable way by leveraging five neural network modules. These modules are trained in an end-to-end manner and each module provides transparent intermediate outputs. Experimental results demonstrate the effectiveness of each module, and analysis on intermediate outputs presents good interpretability for the inference process in contrasted with ``black box" models. Moreover, we find that with explicitly designed compositional modeling of inference process, our approach with a few training examples achieves similar accuracy to strong baselines with full-size training data which indicates a better generalization capability. Meanwhile, extending these models to a larger scope of question types or more complex scenarios is still a challenge, and we will further investigate the trade-off between explainability and scalability.
\section*{Acknowledgement}
We acknowledge this work is supported by National Natural Science Foundation of China (No.61751201) and National Key R\&D Plan (No.2016QY03D0602), We would also like to thank the anonymous reviewers for their insightful suggestions.
\bibliography{emnlp2020.bib}
\bibliographystyle{acl_natbib}

\renewcommand\thesubsection{\Alph{subsection}}

\section*{Appendices}
\label{sec:appendix}

\subsection{Parameters List}
\label{sec:params}
\begin{table}[h!]
\resizebox{\linewidth}{!}{%
\begin{tabular}{@{}lcc@{}}
\toprule
\textbf{Interpretable Reasoning }  &Search Space(Bounds)  & Best Assignment \\ \midrule
No. of GPU(P100)                 & 1                      & 1                \\
Average runtime (mins)     & 45                       &45      \\
No. of params.(include LM)          &361661706                  &361661706\\
No. of Layers in MLP        &3                      &3 \\
No. of Search trials        &38                      &38 \\
learning rate optimizer    & Adam                   &Adam \\
Max Seq. Length            & \textit{choice}[384,512]                     & 512              \\
Doc stride                 & \textit{choice}[64,128]                      & 64               \\
Learning Rate              & \textit{uniform-float}[5e-6,3e-5]                    & 2e-5          \\
Batch Size per GPU         & \textit{choice}[1,2,4]                       & 2                \\
Gradient Accumulation Step & \textit{choice}[1,2]                       & 1                 \\
No. of Epoch               & \textit{uniform-integer}[1,5]                       & 4                 \\
Fixed Length for Q,S,B     & \textit{uniform-integer}{[20,30],[150,250],[350,450]}             & {30,200,400}                  \\ 
\bottomrule
\end{tabular}%
}
\caption{Detailed parameters used in Interpretable Reasoning, we provide  search bounds for each hyperparameter and list out the hyperparameters combination for out best model. Other unmentioned parameters keep same as the one used in BERT.}
\label{tab:params_IR}
\end{table}

\begin{table}[h!]
\resizebox{\linewidth}{!}{%
\begin{tabular}{@{}lcc@{}}
\toprule
\textbf{Answer Prediction}         &Search Space(Bounds)  & Best Assignment \\ \midrule
No. of GPU(P100)                & 4                      & 4                 \\
Average runtime (mins)     & 60                       &60      \\
No. of params.(include LM)          &355361794                  &355361794\\
No. of Search trials        &64                    &64 \\
learning rate optimizer    & Adam                   &Adam \\
Max Seq. Length            & \textit{choice}[384,512]                  & 384              \\
Doc stride                 & \textit{choice}[64,128]                      & 128               \\
Learning Rate              & \textit{uniform-float}[5e-6,3e-5]                    & 1.5e-5            \\
Batch Size per GPU         & \textit{uniform-integer}[4,8]                       & 4                 \\
Answer Length Limit        & \textit{uniform-integer}[5,30]                       & 9                \\
Gradient Accumulation Step & \textit{choice}[1,2]                       & 1                 \\
No. of Epoch               & \textit{uniform-integer}[1,5]                        & 4                 \\
\bottomrule
\end{tabular}%
}
\caption{Detailed parameters used in Answer Prediction, we provide search bounds for each hyperparameter and list out the hyperparameters combination for out best model and baseline model. Other unmentioned parameters keep same as the one used in BERT.}
\label{tab:params_AP}
\end{table}

\subsection{Auxiliary Supervision Instruction}
\label{sec:ASI}
\begin{table}[h!]
{\begin{tabular}[c]{|p{0.95\linewidth}|}
\hline
\textbf{Background}
\\
As a cell grows, its volume increases more quickly than its surface area. If a cell was to get very large, the small surface area would not allow enough nutrients to enter the cell quickly enough for the cell’s needs...Such cell types are found lining your small intestine, where they absorb nutrients from your food through protrusions called microvilli .
\\ 
\textbf{Situation}\\
There are two cells inside a Petri dish in a laboratory, cell X and cell Z. These cells are from the same organism, but are not the same age. Cell X was created two weeks ago, and cell Z was created one month ago. Therefore, cell Z has had two extra weeks of growth compared to cell X. 
\\ 
\textbf{Q\&A}\\ 
Q: Which cell has a larger volume?  \\ A: cell Z\\
\hline
\textbf{Labels} \newline
\begin{tabular}{llc}
World1:&\textit{Cell X}  & [142,148] \\
World2:& \textit{cell Z } & [180,186] \\
Effect:&\textit{volume} &[21,27] \\
$\text{Cause}^{B}$:&\textit{cell grows} &[5,15]\\
$\text{Cause}_{\text{World1}}^{S}$:&\textit{two weeks ago} &[161,174]\\
$\text{Cause}_{\text{World2}}^{S}$:&\textit{one month ago} &[199,212]\\
Cause Cmp: &\textit{cell Z} & 1 \\
Relation: & \textit{positively related} &1\\
Effect Cmp: & \textit{World 2} &1\\
\end{tabular}
\\
\hline
\end{tabular}}
\caption{An example with auxiliary supervision labels.}
\label{tab:label_example}
\end{table}
Table~\ref{tab:label_example} shows one labelled example, and the process of adding auxiliary supervision label contain the following steps:
\begin{enumerate}
    \item Annotate the samples manually: For the selected examples, we find the spans for the worlds, cause property and effect property in the background, cause property for the Worlds in the situation and decide the results for comparison, relation and reasoning modules.
    \item Generate machine-readable labels automatically: Then we use scripts to automatically transform the annotations to the machine-readable form, i.e. we record the start and end character index for all spans and keep the results for comparison, relation and reasoning modules as binary form.
\end{enumerate}

\subsection{Error Cases for Modules}
\label{sec:ECM}
Table~\ref{tab:type1} lists out several type 1 error cases mentioned in the Error Analysis part, while Table~\ref{tab:type2} lists out type 2 error cases which beyonds the scope of our model.

\subsection{More Examples}
\label{sec:more_ex}
We present more examples that correctly answered by our model in Table~\ref{tab:extra_ex}.

\subsection{Heuristic Rules for Answer Prediction}
\label{sec:rules}
We also conduct a rule-based approached to predict the final answer which contains the following steps:
\begin{enumerate}
    \item Categorize questions based on the type of answer: By looking at the labeled train dataset, we can summarize that the types of answer can be divided into two types:1.World Type, answer is one of the compared worlds; 2. Comparative Word Type, like "more" or "less".
    \item For World Type, we filter out such type of questions by searching question keywords, for example, questions started with \{ \textit{What, Which, Who, Where, When}\} usually have world type of answers. Then we determine the results based on the prediction obtained in Reasoning Module.
    \item For Comparative Word Type, we further filter out this type of questions from the rest questions by defining a list of comparative word pair like \{\textit{'more':'less','higher':'lower'...} \}. Then we identify the primary world that being compared in the question and associate it with our identified worlds from Group Detection module, then determine the comparative word for the primary compared world by using the results from Reasoning module.
    \item For the remaining questions, we simply return the world with higher effect property probability from Reasoning module as the final answer. 
\end{enumerate}

\begin{table*}[!h]
\small
\begin{tabular}{p{0.7\linewidth}p{0.28\linewidth}}
\toprule[2pt]
\textbf{Examples}       & \textbf{Prediction}  \\ \midrule
ID:1867731649 \& 350571026 \newline
{\color[HTML]{00009B} Background}: Fish mortality is a parameter used in fisheries population dynamics to account for the loss of fish in a fish stock through death. The mortality can be divided into two types:Natural mortality: the removal of fish from the stock due to causes not associated with fishing. Such causes can include disease, competition, cannibalism, old age, predation, pollution or any other natural factor that causes the death of fish. In fisheries models natural mortality is denoted by (M).[1]Fishing mortality: the removal of fish from the stock due to fishing activities using any fishing gear It is denoted by (F) in fisheries models.
\newline
{\color[HTML]{CE6301} Situation}: Tony is about to go on two fishing trips in the up coming week. On Friday, he is going to Bear Lake, which is located near a factory that has been known to dump waste into the lake. On Saturday, he is going to Fox Lake, which is in a secluded valley.  \newline
{\color[HTML]{009901} Q\&A}:Which day will Tony visit a lake that more likely has more fish? {\color[HTML]{FE0000} \textbf{Saturday}} \newline
{\color[HTML]{009901} Q\&A}:Which lake probably has more fish in it? {\color[HTML]{FE0000} \textbf{Fox Lake}}
&
\textbf{Wrong Predicted Worlds}  \newline
\begin{tabular}{ll}
Worlds: &\textit{[Bear Lake, Fox Lake]} \\
Effect: &\textit{fishing} \\
$\text{Cause}^{B}$: &\textit{Natural mortality}\\
$\text{Cause}_{\text{World1}}^{S}$: &\textit{dump waste}\\ 
$\text{Cause}_{\text{World2}}^{S}$: &\textit{secluded valley} \\ 
Cause Cmp: &\textit{World 1} \\
Relation: &\textit{negatively related }\\
Effect Cmp: &\textit{World 2} \\
\textit{Final Answer}: &Fox Lake
\end{tabular}
\\
\midrule
ID:4215374242 \newline
{\color[HTML]{00009B} Background}: Making these healthy lifestyle choices can also help prevent some types of cancer. In addition, you can lower the risk of cancer by avoiding carcinogens , which are substances that cause cancer. For example, you can reduce your risk of lung cancer by not smoking. You can reduce your risk of skin cancer by using sunscreen. How to choose a sunscreen that offers the most protection is explained below ( Figure below ). Some people think that tanning beds are a safe way to get a tan. This is a myth. Tanning beds expose the skin to UV radiation. Any exposure to UV radiation increases the risk of skin cancer. It doesn't matter whether the radiation comes from tanning lamps or the sun.
\newline
{\color[HTML]{CE6301} Situation}: Steve and Bill are really good friends with each other. The other day they were talking about some habits they have. Steve likes to work out and stay in shape, eats healthy and does not smoke. Bill said he wants to be more like Steve. Bill currently smokes, and loves to go out tanning in the sun without out sunscreen. \newline
{\color[HTML]{009901} Q\&A}:Who has a more likely chance to get lung cancer in the future? {\color[HTML]{FE0000} \textbf{Bill}} 
&
\textbf{Wrong Predicted Comparison Results} \newline
\begin{tabular}{ll}
Worlds: &\textit{[Steve, Bill]} \\
Effect: &\textit{risk of lung cancer} \\
$\text{Cause}^{B}$: &\textit{not smoking.}\\
$\text{Cause}_{\text{World1}}^{S}$: &\textit{does not smoke}\\ 
$\text{Cause}_{\text{World2}}^{S}$: &\textit{smokes} \\ 
Cause Cmp: &\textit{World 2} \\
Relation: &\textit{negatively related }\\
Effect Cmp: &\textit{World 1} \\
\textit{Final Answer}: &Steve
\end{tabular}
\\
\midrule
ID: 4035582237 \newline
{\color[HTML]{00009B} Background}: Sometimes muscles and tendons get injured when a person starts doing an activity before they have warmed up properly. A warm up is a slow increase in the intensity of a physical activity that prepares muscles for an activity. Warming up increases the blood flow to the muscles and increases the heart rate. Warmed-up muscles and tendons are less likely to get injured. For example, before running or playing soccer, a person might jog slowly to warm muscles and increase their heart rate. Even elite athletes need to warm up ( Figure below ).
\newline
{\color[HTML]{CE6301} Situation}: Greg and Carl and about to do a marathon. Greg sees Carl doing some warm ups and laughs to himself and thinks it is silly. They both want to get a good time, and are both avid runners. \newline
{\color[HTML]{009901} Q\&A}:Who is more likely to \textbf{get an injury} during the race? {\color[HTML]{FE0000} \textbf{Greg}} 
&
\textbf{Wrong Predicted Cause Property in Situation}\newline
\begin{tabular}{ll}
Worlds: &\textit{[Greg,Carl]} \\
Effect: &\textit{get injured.} \\
$\text{Cause}^{B}$: &\textit{Warmed-up}\\
$\text{Cause}_{\text{World1}}^{S}$: &\textit{warm ups}\\ 
$\text{Cause}_{\text{World2}}^{S}$: &\textit{warm ups} \\ 
Cause Cmp: &\textit{World 1} \\
Relation: &\textit{negatively related }\\
Effect Cmp: &\textit{World 2} \\
\textit{Final Answer}: &Carl
\end{tabular}
\\
\bottomrule[2pt]
\end{tabular}
\caption{Type 1 error cases made by our model}
\label{tab:type1}
\end{table*}
\begin{table*}[!]
\small
\begin{tabular}{p{0.7\linewidth}p{0.28\linewidth}}
\toprule[2pt]
\textbf{Examples}       & \textbf{Prediction}  \\ \midrule
ID:710693196 \newline
{\color[HTML]{00009B} Background}: Fish reproduce sexually. They lay eggs that can be fertilized either inside or outside of the body. In most fish, the eggs develop outside of the mother's body. In the majority of these species, fertilization also takes place outside the mother's body. The male and female fish release their gametes into the surrounding water, where fertilization occurs. Female fish release very high numbers of eggs to increase the chances of fertilization.
\newline
{\color[HTML]{CE6301} Situation}: All marine creatures are not fish. There are some mammals, for example, whales, also live in the water. Rob wants to know more about differences between fish and other non fish creatures in the water. He divided them into two groups, group A and group B. Group A consists of fish, and group B consists of non fish creatures in the water. He started to see the differences between these two groups \newline
{\color[HTML]{009901} Q\&A}:In group B, would fertilization most likely take place inside or outside of mother's body? {\color[HTML]{FE0000} \textbf{inside}} 
&
\textbf{Missing Knowledge}  \newline
\begin{tabular}{ll}
Worlds: &\textit{[Group A, group B]} \\
Effect: &\textit{fish} \\
$\text{Cause}^{B}$: &\textit{fertilization}\\
$\text{Cause}_{\text{World1}}^{S}$: &\textit{fish}\\ 
$\text{Cause}_{\text{World2}}^{S}$: &\textit{non fish creatures} \\ 
Cause Cmp: &\textit{World 1} \\
Relation: &\textit{positively related }\\
Effect Cmp: &\textit{World 1} \\
\textit{Final Answer}: &Group A
\end{tabular}
\\

\midrule
ID:3339143431 \newline
{\color[HTML]{00009B} Background}: In exercises such as weight lifting, skeletal muscle contracts against a resisting force (see Figure below ). Using skeletal muscle in this way increases its size and strength. In exercises such as running, the cardiac muscle contracts faster and the heart pumps more blood. Using cardiac muscle in this way increases its strength and efficiency. Continued exercise is necessary to maintain bigger, stronger muscles. If you don't use a muscle, it will get smaller and weaker--so use it or lose it.
\newline
{\color[HTML]{CE6301} Situation}: A study was done in the town of Greenwich comparing muscle strength to the amount a person exercises. Mathew goes to the gym 5 times a week and does very intensive workouts. Damen on the other hand does not go to the gym at all and lives a mostly sedentary lifestyle.\newline
{\color[HTML]{009901} Q\&A}:Given Mathew suffers an injury while working out and cannot go to the gym for 3 months, will Mathews strength increase or decrease? {\color[HTML]{FE0000} \textbf{decrease}} 
&
\textbf{Implicit Worlds}\newline
\begin{tabular}{ll}
Worlds: &\textit{[Mathew, Damen]} \\
Effect: &\textit{strength and efficiency} \\
$\text{Cause}^{B}$: &\textit{exercises such as running}\\
$\text{Cause}_{\text{World1}}^{S}$: &\textit{goes to the gym}\\ 
$\text{Cause}_{\text{World2}}^{S}$: &\textit{does not go to the gym} \\ 
Cause Cmp: &\textit{World 1} \\
Relation: &\textit{positively related }\\
Effect Cmp: &\textit{World 1} \\
\textit{Final Answer}: &Mathew
\end{tabular}
\\
\midrule
ID: 2918297602 \newline
{\color[HTML]{00009B} Background}: In common everyday speech, speed of sound refers to the speed of sound waves in air. However, the speed of sound varies from substance to substance: sound travels most slowly in gases; it travels faster in liquids; and faster still in solids. For example, (as noted above), sound travels at 343 m/s in air; it travels at 1,480 m/s in water (4.3 times as fast as in air); and at 5,120 m/s in iron (about 15 times as fast as in air). In an exceptionally stiff material such as diamond, sound travels at 12,000 metres per second (27,000 mph);[1] (about 35 times as fast as in air) which is around the maximum speed that sound will travel under normal conditions.
\newline
{\color[HTML]{CE6301} Situation}: John and Keith are neighbors. They have been pondering about how to communicate with each other in a doomsday scenario when all the electronic devices would be useless. They connected their houses with three ducts. One of the ducts is filled with air; they called it channel A. Another duct is filled with water; they called it channel B. And the last duct is filled with iron; they called it channel C. They can now transmit sound with these channels of communication; in case, disaster strikes \newline
{\color[HTML]{009901} Q\&A}:How much faster would be channel B than channel A in m/s, 1130 m/s or 1137 m/s?{\color[HTML]{FE0000} \textbf{1137 m/s}} 
&
\textbf{Additional Math Computation}\newline
\begin{tabular}{ll}
Worlds: &\textit{[channel A,channel B]} \\
Effect: &\textit{5,120 m/s} \\
$\text{Cause}^{B}$: &\textit{water}\\
$\text{Cause}_{\text{World1}}^{S}$: &\textit{air}\\ 
$\text{Cause}_{\text{World2}}^{S}$: &\textit{water} \\ 
Cause Cmp: &\textit{World 1} \\
Relation: &\textit{positively related }\\
Effect Cmp: &\textit{World 1} \\
\textit{Final Answer}: &channel A
\end{tabular}
\\
\bottomrule[2pt]
\end{tabular}
\caption{Type 2 error cases could not be solved by our model}
\label{tab:type2}
\end{table*}

\begin{table*}[t!]
\small
\begin{tabular}{p{0.7\linewidth}p{0.25\linewidth}}
\toprule[2pt]
\textbf{Examples}       &  \textbf{Prediction}  \\ \midrule
ID:1629973236 \newline
{\color[HTML]{00009B} Background}:One result of air pollution is acid rain. Acid rain is precipitation with a low (acidic) pH. This rain can be very destructive to \textbf{wildlife}. When acid rain falls in forests, freshwater habitats, or soils, it can kill insects and aquatic life. It causes this damage because of its very low pH. Sulfur oxides and nitrogen oxides in the air both cause acid rain to form ( Figure below ). Sulfur oxides are chemicals that are released from coal-fired power plants. Nitrogen oxides are released from motor vehicle exhaust.\newline
{\color[HTML]{CE6301} Situation}: Bill is planning on moving soon. He wants to move to a city that has fresher air and more wildlife to see. His two options that he must choose from are St. Louis and Seattle. Recently,Seattle has installed a new wind farm, and zero emission solar farm to generate power, while St. Louis recently installed a coal fired power plant. Both cities have similar commercial and industrial sectors, and only differ in how they generate power.\newline
{\color[HTML]{009901} Q\&A}: Which city will more likely have more vibrant \textbf{wildlife}? {\color[HTML]{FE0000} \textbf{Seattle}}
&
Module Output: \newline
\begin{tabular}{ll}
Worlds: &\textit{[Seattle,St. Louis]} \\
Effect: &\textit{wildlife} \\
$\text{Cause}^{B}$: &\textit{acid rain.}\\
$\text{Cause}_{\text{World1}}^{S}$: &\textit{wind farm,}\\ 
$\text{Cause}_{\text{World2}}^{S}$: &\textit{coal fired power plant.} \\ 
Cause Cmp: &\textit{World 2} \\
Relation: &\textit{negatively related }\\
Effect Cmp: &\textit{World 1} \\
\textit{Final Answer}: &Seattle
\end{tabular}

\\
\midrule
ID:802123740 \newline
{\color[HTML]{00009B} Background}:One result of air pollution is acid rain. Acid rain is precipitation with a low (acidic) pH. This rain can be very destructive to wildlife. When acid rain falls in forests, freshwater habitats, or soils, it can kill insects and aquatic life. It causes this damage because of its very low pH. Sulfur oxides and nitrogen oxides in the air both cause acid rain to form ( Figure below ). \textbf{Sulfur oxides} are chemicals that are released from coal-fired power plants. Nitrogen oxides are released from motor vehicle exhaust.\newline
{\color[HTML]{CE6301} Situation}: Bill is planning on moving soon. He wants to move to a city that has fresher air and more wildlife to see. His two options that he must choose from are St. Louis and Seattle. Recently, Seattle has installed a new wind farm, and zero emission solar farm to generate power, while St. Louis recently installed a coal fired power plant. Both cities have similar commercial and industrial sectors, and only differ in how they generate power.\newline
{\color[HTML]{009901} Q\&A}: Will Seattle have more or less \textbf{sulfur oxides} in the air than St. Louis? {\color[HTML]{FE0000} \textbf{less}}
&
Module Output: \newline
\begin{tabular}{ll}
Worlds: &\textit{[Seattle,St. Louis]} \\
Effect: &\textit{Sulfur oxides} \\
$\text{Cause}^{B}$: &\textit{coal-fired power plants.}\\
$\text{Cause}_{\text{World1}}^{S}$: &\textit{wind farm,}\\ 
$\text{Cause}_{\text{World2}}^{S}$: &\textit{coal fired power plant.} \\ 
Cause Cmp: &\textit{World 2} \\
Relation: &\textit{positively related }\\
Effect Cmp: &\textit{World 2} \\
\textit{Final Answer}: &St. Louis
\end{tabular}


\\
\midrule
ID:2133492859 \newline
{\color[HTML]{00009B} Background}:Turner et al (2006) derived crash prediction models for this report's predecessor and found a pronounced '2018safety in numbers' effect in the models. Using the crash prediction model for mid-block locations, generic motorist and cyclist volumes can be used to demonstrate the impacts on the expected crash rate of varying motor vehicle and cycle volumes. As shown in figure 2.20, an increase in the proportion of cyclists to the overall traffic volume causes an increase in expected crashes at mid-block locations, but the crash rate increases at a decreasing rate. That is to say, the \textbf{crash rate per cyclist} goes down as the cycle volume increases.\newline
{\color[HTML]{CE6301} Situation}: There were a lot of motorcycles on Interstate 17 last week. On Monday, there were 1355 motorcyclists. On Tuesday, there were 2355 motorcyclists. On Wednesday, there were 3351 motorcyclists. On Thursday, there were 4351 motorcyclists. On Friday, there were 5351 motorcyclists. On Saturday, there were 6351 motorcyclists. On Sunday, there were 7351 motorcyclists. \newline
{\color[HTML]{009901} Q\&A}: What day had a lower \textbf{crash rate per cyclist}: Thursday or Sunday? {\color[HTML]{FE0000} \textbf{Sunday}}
&
Module Output: \newline
\begin{tabular}{ll}
Worlds: &\textit{[Thursday,Sunday]} \\
Effect: &\textit{crash rate per cyclist} \\
$\text{Cause}^{B}$: &\textit{the cycle volume}\\
$\text{Cause}_{\text{World1}}^{S}$: &\textit{4351 motorcyclists}\\ 
$\text{Cause}_{\text{World2}}^{S}$: &\textit{7351 motorcyclists} \\ 
Cause Cmp: &\textit{World 2} \\
Relation: &\textit{negatively related }\\
Effect Cmp: &\textit{World 1} \\
\textit{Final Answer}: &Thursday
\end{tabular}
\\
\midrule
ID:3099625752 \newline
{\color[HTML]{00009B} Background}:The example of someone having a positive experience with a drug is easy to see how drug dependence and the law of effect works. The \textbf{tolerance} for a drug goes up as one continues to use it after having a positive experience with a certain amount the first time. It will take more and more to get that same feeling. This is when the controlled substance in an experiment would have to be modified and the experiment would really begin. The law of work for psychologist B. F. Skinner almost half a century later on the principles of operant conditioning, \"a learning process by which the effect, or consequence, of a response influences the future rate of production of that response.\newline
{\color[HTML]{CE6301} Situation}: The Speed Squad met to discuss their experiences. They all said they always had a great experience using speed and used the same amount each time. They told how many times they used speed. Todd used it 36 times, Jesse used it 40 times, Craig used it 44 times, Alan used it 56 times, Shawn used it 69 times, Clarence used it 78 times, and Sean used it 86 times. \newline
{\color[HTML]{009901} Q\&A}: Who has a higher \textbf{tolerance} for speed: Alan or Clarence? {\color[HTML]{FE0000} \textbf{Clarence}}
&
Module Output: \newline
\begin{tabular}{ll}
Worlds: &\textit{[Alan,Clarence]} \\
Effect: &\textit{tolerance} \\
$\text{Cause}^{B}$: &\textit{positive experience with}\\
$\text{Cause}_{\text{World1}}^{S}$: &\textit{used it 56 times}\\ 
$\text{Cause}_{\text{World2}}^{S}$: &\textit{used it 78 times} \\ 
Cause Cmp: &\textit{World 2} \\
Relation: &\textit{positively related }\\
Effect Cmp: &\textit{World 2} \\
\textit{Final Answer}: &Clarence
\end{tabular}
\\
\bottomrule[2pt]
\end{tabular}
\caption{Examples correctly answered by our model in an intepretable manner.}
\label{tab:extra_ex}
\end{table*}

\end{document}